\documentclass[journal]{IEEEtran}
\usepackage{amsmath}
\usepackage{graphicx}
\usepackage{booktabs}
\usepackage{array}
\usepackage{url}
\usepackage{algorithm}
\usepackage{algorithmic}
\usepackage{tikz}
\usepackage{pgfplots}
\pgfplotsset{compat=1.17}
\usetikzlibrary{shapes.geometric,arrows.meta,positioning,fit,backgrounds}

\begin{document}

\title{A Combined Feature-Based Framework for Disguise and Spoofing Detection in Face Recognition Systems}

\author{S.~Pararajasingham\thanks{Manuscript received August 11, 2026.}\thanks{The author is with the University of Colombo School of Computing (UCSC), University of Colombo, Colombo, Sri Lanka (e-mail: asan935para@gmail.com). ORCID: 0009-0005-3407-2154.}\thanks{A preprint of this manuscript is available at arXiv:2608.08521 [cs.CV], https://arxiv.org/abs/2608.08521.}}

\markboth{IEEE Transactions on Biometrics, Behavior, and Identity Science}{Pararajasingham: Disguise and Spoofing Detection in Face Recognition}

\maketitle

\begin{abstract}
Face recognition systems are increasingly deployed in access-control, mobile, and financial applications, yet remain vulnerable to two distinct classes of attack: spoofing, in which an impostor presents a photograph or video of an authorized user, and disguise, in which a legitimate user is rejected because their current appearance differs from their enrolled template due to accessories, facial hair, cosmetic change, illumination, or pose. Most prior work addresses these two problems separately, using liveness-detection methods for spoofing and occlusion-robust recognition methods for disguise. This paper proposes and empirically compares five combined feature-extraction and classification pipelines that target both problems within a single framework: PM (PCA and Minimum Euclidean Distance, MED), LPM (Local Binary Patterns, LBP, with PCA and MED), HPM (Histogram of Oriented Gradients, HOG, with PCA and MED), SM (Speeded-Up Robust Features, SURF, with MED), and HM (Harris corner features with MED). Each pipeline follows a common two-phase process comprising pre-processing, feature extraction, feature filtering, and classification. The methods were trained on 115 subjects drawn from the FEI, Disguised Faces Database (DFD), and NUAA databases and evaluated on six test conditions covering mixed appearances, frontal faces, dark illumination, left- and right-turned poses, and photo-spoofing attempts. The HOG-based pipeline (HPM) achieved the most consistent performance across conditions, with 94.59\% accuracy on mixed-appearance disguise, 81.5\%--93.2\% across pose and illumination variants, and 91.67\% on spoofing, while the LBP-based pipeline (LPM) achieved the second-highest spoofing-detection accuracy of the five pipelines (93.2\%), behind PM (96.67\%), but weaker robustness to pose change. These results show a measurable trade-off between spoof sensitivity and disguise robustness among classical feature representations, and motivate the deep-learning and cross-database extensions discussed in the concluding sections.
\end{abstract}


\IEEEpeerreviewmaketitle

\section{Introduction}
\IEEEPARstart{A}{utomatic} face recognition is now embedded in mobile-device unlocking, e-commerce, e-health, border control, and mobile-banking applications, largely because it requires no physical token and can operate at a distance without the active cooperation of the subject. This convenience, however, comes with a security cost: a face template is comparatively easy to acquire without contact, which exposes recognition systems to two related but distinct failure modes.

The first, spoofing, occurs when an unauthorized individual presents a photograph, replayed video, or other counterfeit likeness of an authorized user in order to obtain that user's access rights \cite{kose2014}. The most common spoofing attacks against 2D systems are carried out with printed photographs and simple video replay, precisely because these require no specialised equipment and can be produced from images gathered from social media or public records. The second failure mode, disguise, occurs when a legitimate, enrolled user is rejected because their presented appearance has changed relative to the enrolled reference image, whether through eyewear, facial hair, cosmetic change, hairstyle, illumination, pose, or long-term ageing. Both failure modes degrade the practical reliability of face recognition, yet they are commonly studied and countered separately: liveness- and texture-based anti-spoofing methods on one hand, and occlusion- or pose-robust recognition methods on the other.

This paper addresses both problems within a single evaluation framework rather than proposing one new descriptor. Four established feature-extraction techniques -- Local Binary Patterns (LBP), Histogram of Oriented Gradients (HOG), Speeded-Up Robust Features (SURF), and the Harris corner detector -- are combined with two established filtering strategies (Principal Component Analysis, PCA, and strongest-keypoint selection) and a common Minimum Euclidean Distance (MED) classifier. The resulting five pipelines are evaluated side by side, under identical training and test conditions, so that the accuracy trade-offs between spoof sensitivity and disguise robustness can be measured directly rather than inferred from separate studies that use different protocols.

The contributions of this paper are threefold: (i) a unified two-phase pipeline that treats spoof detection and disguise-robust recognition as a single classification decision rather than two separate subsystems; (ii) a systematic, identical-protocol comparison of five classical feature-extraction and classification combinations across six real-world test conditions; and (iii) an explicit accuracy breakdown by attack and variation type that identifies where each combination is strong and where it fails, used in Section~V to motivate directions for follow-up work. Portions of the methodology and evaluation reported here extend an earlier unpublished MCS dissertation on the same topic \cite{pararajasingham2019}.

The remainder of this paper is organized as follows. Section~II reviews related work; Section~III describes the datasets and methodology; Section~IV reports results; Section~V discusses findings and limitations; Sections~VI and VII conclude and outline future work.

\section{Related Work}

\subsection{Spoofing Attacks and Countermeasures}
Motion-based countermeasures analyse subconscious liveness cues such as eye blinking and head or mouth movement \cite{kollreider2007}. Because such cues occur at a physiological rate of roughly 0.2--0.5~Hz, these methods typically require several seconds of video to accumulate a stable decision \cite{kollreider2007}; eye-blink-specific detectors exploit the fact that a live subject blinks approximately every two to four seconds, a cue that is difficult to reproduce convincingly with a static photograph or simple replay \cite{pan2007}. Texture-based countermeasures instead operate on a single image, analysing the frequency-domain or micro-texture differences between a live face and a recaptured photograph or screen replay \cite{li2004}, exploiting the observation that recaptured and genuinely captured images retain subtle but detectable similarities and differences \cite{tan2010}. Local Binary Pattern Variance (LBPV) is one such descriptor, jointly encoding local texture pattern and contrast information from sampling points around a circular neighbourhood \cite{guo2010}; more broadly, LBP-, DoG-, and HOG-based texture descriptors have been shown to discriminate recapture artefacts from genuine skin texture using only a single frame \cite{yang2013}. Image-quality-analysis methods generalise this idea further, designing features that are explicitly sensitive to the quality loss introduced by print or replay, and have reported strong within-database performance on the Idiap REPLAY-ATTACK and CASIA-FASD benchmarks \cite{galbally2014,wen2015}, with public benchmarking competitions on these datasets providing a further basis for cross-method comparison \cite{chingovska2013}. Methods that fuse additional sensing modalities, such as 3D depth, infra-red, or contextual scene cues, can achieve high robustness but impose extra hardware or protocol requirements, and specific contextual cues have been shown to be circumventable once the spoofing mechanism is known \cite{komulainen2013}.

Table~\ref{tab:antispoofing} summarises these four broad categories of spoofing countermeasure with their principal strengths, limitations, and representative reported performance.

\begin{table*}[t]
\caption{Comparison of Face Anti-Spoofing Method Categories Reported in the Reviewed Literature}
\label{tab:antispoofing}
\centering
\begin{tabular}{p{3.2cm}p{3.0cm}p{4.3cm}p{4.7cm}}
\toprule
\textbf{Method Category} & \textbf{Representative Cue} & \textbf{Strengths} & \textbf{Limitations} \\
\midrule
Motion-based \cite{kollreider2007,pan2007,sun2007,bao2009} & Eye blink, head/mouth movement & Good generalisation across capture devices & Low robustness to replayed motion; slow response ($>$3~s); high computational cost \\
Texture-based \cite{guo2010,yang2013,li2004} & LBP / DoG / HOG micro-texture & Fast response ($<$1~s); low computational cost & Sensitive to acquisition-condition changes; weaker cross-database generalisation \\
Other cues \cite{komulainen2013,hayat2016,morency2003} & 3D depth, IR, context, voice & High robustness within protocol & Needs additional sensing hardware; slow response with audio/3D cues \\
Image-quality analysis \cite{galbally2014} & Recapture-induced quality loss & Good generalisation; fast response; low cost & Often needs different classifiers tuned per attack type \\
\bottomrule
\end{tabular}
\end{table*}

\subsection{Disguise Variations in Face Recognition}
Disguise-robust recognition is complicated by low-quality or non-cooperative capture, temporal appearance change, and deliberate occlusion by accessories. Eigen-space approaches restricted to the periocular region (``eigen-eyes'') have been used to reduce sensitivity to occlusion below the eyes, reporting 87.5\% accuracy on the Yale face database \cite{morency2003,georghiades2001}, while feature-fusion approaches combining RGB and depth data from Kinect sensors have reported improved robustness to pose and partial occlusion relative to 2D-only baselines \cite{hayat2016}. Benchmarking on the AR Face Database, which contains variation in eyewear, expression, and illumination, shows that even specialised occlusion-handling algorithms remain sensitive to the specific accessory and illumination combination presented \cite{kose2014,martinez1998}; a nearest-neighbour classifier using skin-correlation features, for example, achieved only 45.8\% accuracy on this benchmark. Beyond accessory occlusion, facial similarity has also been shown to degrade jointly across age progression, disguise, illumination, and pose \cite{ramanathan2004}, underscoring that disguise-robust recognition must contend with several confounding variation types simultaneously rather than in isolation. This body of work motivates evaluating disguise robustness across explicit, separated variation types rather than reporting a single pooled accuracy figure, the approach adopted in Section~IV.

\section{Materials and Methods}

\subsection{Datasets}
Three public face databases were used. FEI \cite{fei2019} provides pose and expression variation for disguise evaluation; the Disguised Faces Database (DFD) provides accessory, facial-hair, and cosmetic variation; and NUAA provides photo-spoofing attempts. Table~\ref{tab:datasets} summarises the full composition of each source database. The training set combined 115 identity subjects (50 from FEI, 15 from NUAA, 50 from DFD)\footnote{Verified against the original project files; corrects a per-database breakdown in the earlier dissertation \cite{pararajasingham2019} (15 from FEI, 50 from NUAA). NUAA contains only 15 subjects in total. Reported accuracy results are unaffected.}, used jointly by a single trained model for both the disguise-recognition and spoofing-detection tasks, evaluated identically via the classification rules in Section~III-F. The corresponding test set was partitioned into six condition-specific subsets: mixed appearances (37 images), frontal faces (200), left-turned faces (250), right-turned faces (250), dark-illumination faces (50), and spoofing attempts (60, from NUAA).

\begin{table}[t]
\caption{Databases Used for Training and Testing}
\label{tab:datasets}
\centering
\begin{tabular}{lccc}
\toprule
\textbf{Database} & \textbf{Subjects} & \textbf{Img./Subj.} & \textbf{Total} \\
\midrule
FEI  & 200 & 14 & 2{,}800 \\
DFD  & 409 & $\sim$6 & 2{,}460 \\
NUAA & 15  & -- & 10{,}230 \\
\bottomrule
\end{tabular}

\vspace{4pt}
\begin{minipage}{\columnwidth}
\scriptsize DFD: 6 images/subject is a nominal average; the per-subject count varies slightly. NUAA is organised as paired genuine/imposter capture sessions (5{,}115 each) rather than a fixed number of images per subject.
\end{minipage}
\end{table}

\begin{table}[t]
\caption{Combined Pipelines and Decision Thresholds}
\label{tab:pipelines}
\centering
\begin{tabular}{lllc}
\toprule
\textbf{Pipeline} & \textbf{Feature} & \textbf{Decision Basis} & $\boldsymbol{\tau}$ \\
\midrule
PM  & --    & Min. Euclidean dist. & $6.50\times10^{17}$ \\
LPM & LBP   & Min. Euclidean dist. & $7.60\times10^{17}$ \\
HPM & HOG   & Min. Euclidean dist. & $3.30\times10^{5}$ \\
SM  & SURF  & Matched keypoints    & $20$ \\
HM  & Harris& Matched keypoints    & $4$ \\
\bottomrule
\end{tabular}
\end{table}

\subsection{System Overview}
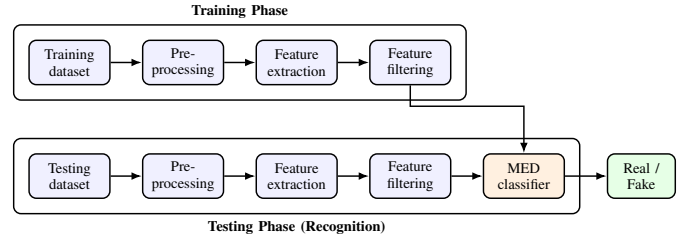
\begin{figure}[t]
\centering
\resizebox{\linewidth}{!}{%
\begin{tikzpicture}[
  font=\footnotesize,
  stage/.style={draw, rounded corners, minimum width=1.55cm, minimum height=0.85cm, align=center, fill=blue!6, thick},
  phase/.style={draw, thick, rounded corners, inner sep=8pt},
  arr/.style={-{Latex[length=2mm]}, thick}
]
 
 \node[stage] (tr-in)  {Training\\dataset};
 \node[stage, right=6mm of tr-in]  (tr-pp)  {Pre-\\processing};
 \node[stage, right=6mm of tr-pp]  (tr-fe)  {Feature\\extraction};
 \node[stage, right=6mm of tr-fe]  (tr-ff)  {Feature\\filtering};
 \draw[arr] (tr-in) -- (tr-pp);
 \draw[arr] (tr-pp) -- (tr-fe);
 \draw[arr] (tr-fe) -- (tr-ff);
 \begin{scope}[on background layer]
   \node[phase, fit=(tr-in)(tr-pp)(tr-fe)(tr-ff), label=above:{\bfseries Training Phase}] (trphase) {};
 \end{scope}

 \node[stage, below=13mm of tr-in]  (te-in)  {Testing\\dataset};
 \node[stage, right=6mm of te-in]  (te-pp)  {Pre-\\processing};
 \node[stage, right=6mm of te-pp]  (te-fe)  {Feature\\extraction};
 \node[stage, right=6mm of te-fe]  (te-ff)  {Feature\\filtering};
 \node[stage, right=6mm of te-ff, fill=orange!12]  (te-cl)  {MED\\classifier};
 \draw[arr] (te-in) -- (te-pp);
 \draw[arr] (te-pp) -- (te-fe);
 \draw[arr] (te-fe) -- (te-ff);
 \draw[arr] (te-ff) -- (te-cl);
 \begin{scope}[on background layer]
   \node[phase, fit=(te-in)(te-pp)(te-fe)(te-ff)(te-cl), label=below:{\bfseries Testing Phase (Recognition)}] (tephase) {};
 \end{scope}

 \draw[arr] (tr-ff.south) -- ++(0,-4mm) -| (te-cl.north);

 \node[stage, right=8mm of te-cl, fill=green!10, minimum width=1.3cm] (out) {Real /\\Fake};
 \draw[arr] (te-cl) -- (out);

\end{tikzpicture}
}
\caption{Overall two-phase process. Training-phase filtered features (PCA eigenspace or strongest keypoints) supply the reference model used by the MED classifier in the testing phase.}
\label{fig:pipeline}
\end{figure}

The proposed framework follows a common two-phase process, training and testing, each composed of the same four stages: pre-processing, feature extraction, feature filtering, and classification. Five combined pipelines are instantiated from this common process by varying the feature-extraction and filtering stage: PM (PCA + MED), LPM (LBP + PCA + MED), HPM (HOG + PCA + MED), SM (SURF + MED), and HM (Harris + MED). Fig.~\ref{fig:pipeline} illustrates the overall process.

\subsection{Pre-processing}
Input images are $640\times480$ RGB colour images. Each image is cropped to a fixed $280\times380$ region, with crop origin $(180,40)$, then converted from RGB to greyscale prior to feature extraction.

\subsection{Feature Extraction}
LBP, HOG, SURF, and Harris corner features are widely used and complementary local descriptors for face recognition \cite{priyanka2015}. LBP characterises local texture by comparing each pixel to its circular neighbourhood, encoding the result as a rotation-sensitive binary pattern \cite{yang2013}. HOG encodes local edge orientation and is comparatively insensitive to uniform brightness change. SURF is a scale- and in-plane-rotation-invariant detector/descriptor pair using a Hessian-matrix approximation and a 64-dimensional Haar-wavelet-based descriptor \cite{rublee2011}. The Harris--Stephens detector locates corner-like interest points from the directional variation of a local autocorrelation function \cite{harris}. Fig.~\ref{fig:features} summarises what each descriptor responds to and the dimensionality it produces prior to filtering.

\begin{figure}[t]
\centering
\begin{tikzpicture}[
  font=\scriptsize,
  desc/.style={draw, thick, rounded corners, minimum width=3.9cm, minimum height=0.9cm, align=left, inner sep=4pt},
]
\node[desc, fill=blue!8]   (lbp)  {\textbf{LBP} --- pixelwise circular binary pattern; captures local micro-texture};
\node[desc, fill=orange!10, below=3mm of lbp]  (hog)  {\textbf{HOG} --- 8-neighbour gradient direction; captures edge/shape structure};
\node[desc, fill=green!10, below=3mm of hog]  (surf) {\textbf{SURF} --- Hessian-based keypoints; 64-D descriptor, scale/rotation invariant};
\node[desc, fill=red!8, below=3mm of surf] (harris) {\textbf{Harris} --- autocorrelation-based corner keypoints};
\end{tikzpicture}
\caption{Summary of the four feature-extraction descriptors used by the LPM, HPM, SM, and HM pipelines respectively (PM uses raw pixel intensity with no descriptor).}
\label{fig:features}
\end{figure}

\subsection{Feature Filtering}
For LPM and HPM, PCA is applied to the corresponding feature vectors. Each training image is reshaped into a column vector $x_i$, and the mean face vector is
\begin{equation}
\bar{x} = \frac{1}{n}\sum_{i=1}^{n} x_i .
\end{equation}
Each image is mean-centred, $a_i = x_i - \bar{x}$, and combined into $A = [a_1,\dots,a_n]$. Rather than diagonalising the large $MN\times MN$ covariance matrix $C = AA^{\mathsf T}$, the eigenvectors of the smaller $n\times n$ matrix
\begin{equation}
L = A^{\mathsf T}A
\end{equation}
are computed via singular value decomposition, and the eigenvectors of $C$ recovered from those of $L$. The eigenvectors with the largest eigenvalues (the ``eigenfaces'') are retained. For SM and HM, PCA is not applied; the built-in strongest-point selection retains a fixed number of the most salient SURF or Harris keypoints instead.

\subsection{Classification}
The MED classifier is used uniformly, with two decision rules. For PM, LPM, HPM, the minimum Euclidean distance between the test image's projected weight vector $w_{\text{test}}$ and each class mean $w_i$ is
\begin{equation}
d_{\min} = \min_i \lVert w_{\text{test}} - w_i \rVert^2 ,
\end{equation}
and the input is classified as fake if $d_{\min} > \tau$, otherwise matched to $\arg\min_i \lVert w_{\text{test}} - w_i\rVert^2$. For SM and HM, the number of matched keypoints $m$ between the test image and the best-matching candidate is compared to $\tau$: the input is matched if $m > \tau$, otherwise classified as fake (see Table~\ref{tab:pipelines}).

\begin{table*}[!t]
\caption{Recognition Accuracy (\%) by Test Condition}
\label{tab:results}
\centering
\begin{tabular}{lcccccc}
\toprule
\textbf{Pipeline} & \textbf{Mixed (n=37)} & \textbf{Front (n=200)} & \textbf{Left (n=250)} & \textbf{Right (n=250)} & \textbf{Dark (n=50)} & \textbf{Spoof (n=60)} \\
\midrule
PM (PCA+MED)       & 70.27 & 76.50 & 84.40 & 77.60 & 38.00 & \textbf{96.67} \\
LPM (LBP+PCA+MED)  & 75.68 & 66.50 & 56.40 & 57.20 & 72.00 & 93.20 \\
HPM (HOG+PCA+MED)  & \textbf{94.59} & 81.50 & \textbf{91.60} & \textbf{93.20} & \textbf{86.00} & 91.67 \\
SM (SURF+MED)      & 81.08 & 78.50 & 36.40 & 37.20 & 26.00 & 86.67 \\
HM (Harris+MED)    & 81.08 & \textbf{94.50} & 64.80 & 64.80 & 68.00 & 70.00 \\
\bottomrule
\end{tabular}
\end{table*}

\begin{figure*}[!t]
\centering
\begin{tikzpicture}
\begin{axis}[
  width=\textwidth, height=5.2cm,
  ybar, bar width=4.2pt,
  ymin=0, ymax=100,
  ylabel={Accuracy (\%)},
  symbolic x coords={Mixed,Front,Left-turn,Right-turn,Dark,Spoofing},
  xtick=data,
  legend style={at={(0.5,-0.28)}, anchor=north, legend columns=5, font=\scriptsize, draw=none},
  ymajorgrids=true, grid style={dashed, gray!30},
  tick label style={font=\scriptsize},
  label style={font=\scriptsize},
  enlarge x limits=0.09,
  axis line style={draw=gray!60},
]
\addplot[fill=blue!55!black]   coordinates {(Mixed,70.27) (Front,76.50) (Left-turn,84.40) (Right-turn,77.60) (Dark,38.00) (Spoofing,96.67)};
\addplot[fill=orange!85!black] coordinates {(Mixed,75.68) (Front,66.50) (Left-turn,56.40) (Right-turn,57.20) (Dark,72.00) (Spoofing,93.20)};
\addplot[fill=green!55!black]  coordinates {(Mixed,94.59) (Front,81.50) (Left-turn,91.60) (Right-turn,93.20) (Dark,86.00) (Spoofing,91.67)};
\addplot[fill=red!70!black]    coordinates {(Mixed,81.08) (Front,78.50) (Left-turn,36.40) (Right-turn,37.20) (Dark,26.00) (Spoofing,86.67)};
\addplot[fill=violet!70!black] coordinates {(Mixed,81.08) (Front,94.50) (Left-turn,64.80) (Right-turn,64.80) (Dark,68.00) (Spoofing,70.00)};
\legend{PM,LPM,HPM,SM,HM}
\end{axis}
\end{tikzpicture}
\caption{Recognition accuracy (\%) of the five pipelines across the six test conditions (data from Table~\ref{tab:results}). HPM (green) is the most consistently high performer outside spoofing, where PM (blue) and LPM (orange) lead.}
\label{fig:barchart}
\end{figure*}
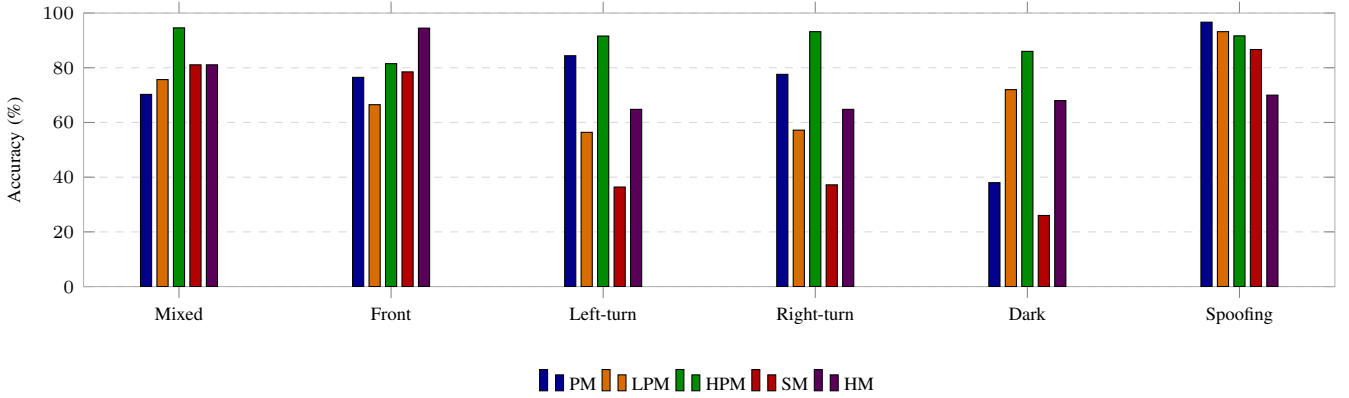

\section{Results}
Table~\ref{tab:results} reports the recognition accuracy achieved by each pipeline across the six test conditions. HPM achieves the highest accuracy on four of six conditions -- mixed-appearance disguise (94.59\%), left-turned (91.6\%), right-turned (93.2\%), and dark illumination (86.0\%) -- and a competitive 91.67\% on spoofing. PM achieves the single highest spoofing accuracy (96.67\%) but the lowest dark-illumination accuracy (38.0\%). LPM achieves the second-highest spoofing accuracy (93.2\%) but falls to 56.4\%/57.2\% on left-/right-turned faces. SM and HM are competitive on spoofing (86.67\%/70.0\%) and frontal faces (78.5\%/94.5\%) but degrade sharply under pose change and dark illumination. Fig.~\ref{fig:barchart} visualises these results, making the spoof-versus-disguise trade-off discussed in Section~V directly visible: the blue and orange bars (PM, LPM) peak on the rightmost group (spoofing) but dip lowest on the dark-illumination group, while the green bars (HPM) stay comparatively high across every group except a small gap on spoofing.

\section{Discussion}
Two patterns emerge from Table~\ref{tab:results}. First, HPM is the most consistently strong pipeline, plausibly because HOG's gradient-orientation encoding is comparatively insensitive to illumination change. Second, no pipeline strong on spoofing is simultaneously strongest on pose variation: PM and LPM's global PCA representation suits detecting a flat photographic replay, while HPM's gradient encoding suits the fine-grained changes of disguise; the keypoint-based pipelines (SM, HM) underperform on pose and illumination because local keypoints are sensitive to out-of-plane rotation and low contrast.

These figures are a same-protocol relative comparison, not directly comparable to headline HTER/EER metrics on Idiap REPLAY-ATTACK or CASIA-FASD, which use different databases and protocols.

\subsection{Limitations and Threats to Validity}
Several limitations apply. First, decision thresholds were fixed from the training set, not calibrated on a held-out set. Second, training and testing used the same combined databases, so results are within-protocol rather than cross-database. Third, evaluated attacks are limited to 2D photographic spoofing; video-replay and 3D-mask attacks are out of scope. Fourth, all pipelines use hand-crafted descriptors, not compared against CNN- or transformer-based baselines, despite neural-network-based approaches being explored since the early 2010s \cite{verma2012}. Fifth, the training population (115 subjects) is modest by contemporary standards.

\section{Conclusion}
This paper evaluated five combined feature-extraction and classification pipelines -- PM, LPM, HPM, SM, and HM -- for the joint problem of disguise-robust recognition and photo-spoofing detection in 2D face recognition. Under an identical protocol across three public databases and six explicit test conditions, HPM provided the most consistent performance across pose, illumination, and disguise variation (81.5\%--94.59\%) while remaining competitive on spoofing (91.67\%), whereas PM and LPM achieved higher peak spoofing accuracy (96.67\% and 93.2\%) at the cost of reduced robustness to pose and illumination change.

\section{Future Work}
Four directions follow from the limitations in Section~V-A: (i)~cross-database evaluation; (ii)~extension to video-replay and 3D-mask spoofing with temporal liveness cues; (iii)~direct comparison against CNN- or transformer-based baselines under the same protocol; and (iv)~explicit modelling of long-term appearance change (ageing).

\section*{Acknowledgment}
The author would like to thank Prof.\ N.~D.~Kodikara for his supervision and guidance during the original MCS dissertation research at the University of Colombo School of Computing, upon which this work is based.

\section*{Declarations}
\textbf{Funding:} This research received no specific grant from any funding agency in the public, commercial, or not-for-profit sectors.
\textbf{Conflict of Interest:} The author declares no known competing financial interests or personal relationships that could have appeared to influence this work.
\textbf{Data Availability:} The datasets used (FEI, DFD, NUAA) are publicly available from their respective sources, cited in the References.
\textbf{Ethical Approval:} This study used only publicly available, previously de-identified face image databases; no new human-subject data were collected.

\end{document}